# Knowledge Representation for Conceptual, Motivational, and Affective Processes in Natural Language Communication

Seng-Beng Ho
Deptartment of Social and Cognitive Computing
Institute of High Performance Computing
Singapore
hosb@ihpc.a-star.edu.sg

Zhaoxia Wang
School of Computing and Information Systems
Singapore Management University
Singapore
zxwang@smu.edu.sg

Boon-Kiat Quek
Department of Social and Cognitive Computing
Institute of High Performance Computing
Singapore
quekbk@ihpc.a-star.edu.sg

Erik Cambria
School of Computer Science and Engineering
Nanyang Technological University
Singapore
cambria@ntu.edu.sg

***Abstract*—Natural language communication is an intricate and complex process. The speaker usually begins with an intention and motivation of what is to be communicated, and what effects are expected from the communication, while taking into consideration the listener's mental model to concoct an appropriate sentence. The listener likewise has to interpret what the speaker means, and respond accordingly, also with the speaker's mental state in mind. To do this successfully, conceptual, motivational, and affective processes have to be represented appropriately to drive the language generation and understanding processes. Language processing has succeeded well with the big data approach in applications such as chatbots and machine translation. However, in human-robot collaborative social communication and in using natural language for delivering precise instructions to robots, a deeper representation of the conceptual, motivational, and affective processes is needed. This paper capitalizes on the UGALRS (Unified General Autonomous and Language Reasoning System) framework and the CD+ (Conceptual Dependency Plus) representational scheme to illustrate how social communication through language is supported by a knowledge representational scheme that handles conceptual, motivational, and affective processes in a deep and general way. Though a small set of concepts, motivations, and emotions is treated in this paper, its main contribution is in articulating a general framework of knowledge representation and processing to link these aspects together in serving the purpose of natural language communication for an intelligent system.**

## I. Introduction

In current AI research, natural language understanding is typically treated separately from natural language generation, and also natural language understanding is typically processed at a superficial level, with no deep representation of meaning [1], [2]. Despite the fact that there is typically no deep meaning representation present in these language processing systems, conversational systems such as chatbots and various machine translation systems have achieved quite significant degrees of commercial successes [3].

However, there are certain applications of natural language understanding that require a deeper or grounded level of representations. For example, in using natural language instructions to instruct a robot to carry out certain actions, the robotic system involved has to comprehend the instructions to a level at which the grounded or "true" meanings are recovered, represented, and put to use in the form of actions and behaviors. Also, motivational and affective processes feature prominently in human language communication [4]. For language generation, firstly there has to be some motivations involved, modulated by the ongoing emotional states, before an utterance is constructed and emitted. For language understanding, a model of the utterer's motivational and emotional states has to be present in the listener's "mind," namely the internal language understanding processes, before a sentence spoken can be appropriately understood with the attendant subsequent correct responses by the listener. The understanding of sentiments present in sentences has been an extensively researched area in AI in the form of sentiment analysis [5]–[11]. However, there has been no integrated natural language communication framework proposed that links language generation and understanding processes together in a principled manner, at the same time incorporating a principled treatment of motivational and emotional processes which are intimately connected with language generation and understanding [1], [2]. This paper attempts to address these issues using the framework of the UGALRS architecture proposed by Ho [12]. The remainder of this paper is organized as follows: Section II discusses the motivation and background; Section III presents the basic architectural and representational constructs; Section IV describes the core of the representations for language communication; finally, Section V proposes concluding remarks.

## II. Motivation and Background

Natural language communication is a two-way process – generation and understanding. When a human emits an utterance, it is with a listener in mind, and it is also at the end of a long process of beginning with a motivation and intention to communicate about something, and then, depending on the context and emotional state of the human involved and the intended effect on the listener, an appropriate sentence is then concocted to precisely communicate the information to the intended listener [29]. To do this successfully, the utterer must also have a good mental model of the mental and emotional states of the listener.

Current big data approaches to language processing, such as the approach taken by the GPT-3 model [3], though successful in many applications, do not model these internal processes. The downside of that is that certain explanations and subsequent related corrective actions or suggestions are not possible. For example, suppose Person A asks Person B, “Why do you say that to him?” [II.1] And Person B replies, “I want him to hurt.” [II.2] Person A may then suggest, “Well, I would suggest a better way to do that, which is to…” [III.3]. Utterance II.2 is an explanation of why Person B said something, and Person B, having a model of her own mental processes beginning with her motivation for uttering the earlier sentence Person A is asking her about, is able to give Person A an *explanation* of what “causes” her to say the earlier thing that was intended to hurt a certain person. Person A, who presumably also has an internal model of herself, Person B, and perhaps also the person whom Person B wants to hurt, is then able to reason about and understand the various mental causalities involved and propose a different way to generate hurt to the person Person B intends to hurt. Whether her suggestion will be a malicious one intended to go along with and please Person B or a benign one to placate the situation for the betterment of all involved will depend on the background of their conversation and other intentions of Person A to start with. Processes such as these are present when humans communicate with robots to instruct them to carry out certain tasks, or when robots are communicating with each other in collaborating to perform certain tasks, and a robotic or AI system would benefit from a fuller model of the internal processes of language generation, communication, and understanding.

A robotic system that is involved in language communication is basically an autonomous system, and language generation and understanding processes involve a number of components in an autonomous system. Robotic systems that are in communication and collaboration with other systems are social agents. Therefore, to fully elucidate the processes involved in language communication, it will be beneficial to look at related work in which an internal operating architecture of an autonomous system or social agent is proposed. Also, the issues of grounded language representation need to be addressed. Ho [12] provides an autonomous system architecture, the Unified General Autonomous and Language Reasoning System (UGALRS), that elucidates various essential operating components, as well as a grounded meaning representation framework based on an enhanced version of the conceptual dependency (CD) theory [13]–[15] called CD+, that operates within UGALRS. In this paper, we will use UGALRS and CD+ as our representational framework.

As elucidated earlier, motivational and affective processes are intimately involved in language communication processes, which have not been adequately addressed in most language-related work, whether they are work in linguistics [16]–[19] or computational linguistics and natural language processing [1], [2], [20]. As will be illustrated, the UGALRS plus CD+ framework provides the necessary representational constructs to handle motivational and affective processes as well. Motivational and affective processes are essential components in the functioning of a social agent involved in communication and collaboration, thus, UGALRS is also an architecture for a social agent.

Quek [21]–[23] has articulated an architecture that incorporates motivation and emotion in the functioning of an autonomous system. Though his work is directed toward the usual robotic actions, driven by motivational and affective processes, it dovetails with the current work in which the “actions” involved would correspond to language generation in the communication process.

As each of the domains of conceptual, motivational, and affective processes is extensive in itself, in this paper we focus on using a small subset of each aspect to articulate and elucidate the intricate connections between them in a general framework that can be extended in future work.

## III. Basic Architectural and Representational Constructs

As mentioned above, our framework is based on the UGALRS architecture and the associated CD+ representational scheme. In this section, we will discuss them accordingly.

### *A. The CD+ Concept Representational Framework*

Ho [12] developed a general representational framework that can be used to represent a large variety, if not all, concepts. It is highlighted in [12] that many of the often-encountered concepts are functional in nature. The framework is termed CD+, which is in turn derived from Schank’s CD conceptual representational framework [13]–[15]. The representations used in CD+ are cognitive, causal, and grounded.

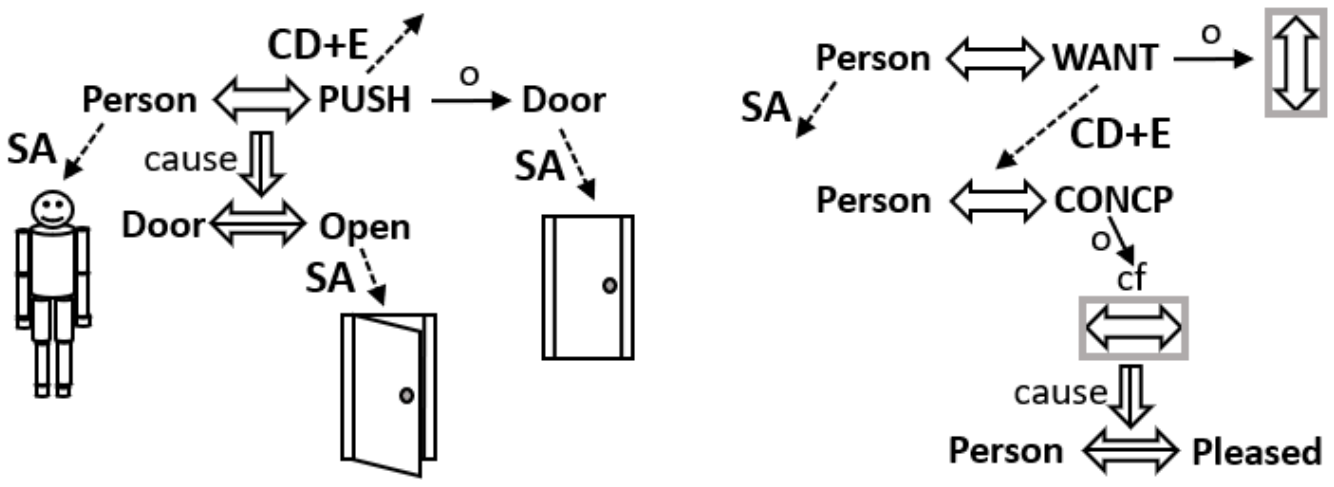

(a) (b)

Fig. 1. Examples of CD+ representations. (a) “Person pushes the door open.” (b) Person wants something.

Fig. 1(a) shows the representation of the sentence, or *conceptualization*, “Person pushes the door open.” The horizontal double arrow links the subject, Person, to the *object* (o), Door. This is termed a conceptualization. In CD+, the two main constructs being enhanced over CD are the Structure Anchor (SA) and the CD+ Elaboration (CD+E) [12]. SA is a detailed structural or analogical representation of the object the symbol refers to, which provides grounding for the symbol. In this case, for a Person, there would be the representational details of the various parts and the overall structure, specified to the details of every point, limbs, and joints, as shown. A convenient way to implement this representational model would be the kind of representation used for objects in computer graphics, which is an analogical representation in the form of a high density point cloud consisting of points corresponding to every point of the object involved, or some vectorial representations that represent the loci of these points. The various parts on the object involved, say, in the case of Person, the various body parts such as the head, torso, and limbs, may be movable relative to each other, and these have to be captured in the model. CD+E is used to elaborate on the symbols that represent certain actions, such as in this case, PUSH. PUSH involves a number of sub-steps, such as “first place palm flat on Door near center of Door,” “then exert strength in the direction perpendicular to Door’s surface,” etc. Here we use English sentences to describe these steps for the ease of illustration, but each of these steps is in turn representable in CD+ form. There is a hierarchy of details until a “ground level” is reached in which the concepts used are “ground level concepts.” A number of ground level concepts is listed in [12] that is supposed to be the common ground for all (or at least a large majority of) concepts. In the two English sentences above used to describe the CD+E involved, every concept used in their representations has to be clearly defined, in further elaboration in the form of CD+E, or in the form of a ground level concept (which could be a SA). So, all the symbols, “first,” “place,” “palm,” “near,” “center,” “of,” and “Door” in the sentence “first place palm near center of Door” have to be defined and the CD+ framework that makes this possible is described in detail in [12]. For the rest of the paper, we will omit the details of the SAs and some of the CD+Es to avoid clutter. The use of the SAs and some of the CD+Es are more relevant to other tasks as discussed in [12].

The sentence “Person pushes the door open” connotes some causality that is not explicit in the sentence. The implicit concept involved is “Person pushes the door and it *causes* the door to open.” Therefore, in Fig. 1(a), there is a vertical arrow with a line down in the middle representing the causality involved. The horizontal double arrow with a line down in the middle represents the “state” of something, in this case, the *state* of Door is Open after being pushed. The concept of Open also needs to be grounded, in this case in the form of an SA.

In Fig. 1(b), a more complex conceptualization is shown. It involves the concept of WANT. In a sentence such as “Person wants X,” in which X could be an object (say, “ice cream”) or it could be another conceptualization (say, “the house to be demolished”), the hidden causality is that “if X is obtained or can be realized, Person will be pleased.” (The *object* of WANT is indicated with a link labelled with an “o”) “Pleased” is a fundamental ground level concept capturing the emotional state of the person involved. It is considered a basic emotional state.

In Fig. 1(b), the vertical double arrow with a gray box around it represents any conceptualization (such as “House be Demolished”), and when Person “WANT” that, the concept of WANT is CD+ Elaborated into a *causal* connection between, say, “House be Demolished” (the same conceptualization of the object of WANT) and Person being in the state of Pleased. The “c” above the horizontal gray box represents the conditional (“if”), and “f” represents the future tense. I.e., “*if* the house is demolished, it *will* cause Person to be Pleased.” Now, this entire causation is in turn a conceptualization created in Person’s internal mental processes.

Therefore, this entire conceptualization is the object of what Person conceptualizes - CONCP (CONCePtualize). There are other more complex CD+ constructs discussed in [12] but these examples would suffice for our subsequent discussion.

## *B. The UGALRS Architecture*

As mentioned above, CD+ representations operate within a general autonomous system (or social agent) architecture, the UGALRS, for the representation of concepts [12]. Fig. 2 shows part of the full UGALRS architecture that focuses on the language aspects. The focus of our attention is on the LANGUAGE COMMUNICATION REASONING CORE (LCRC) but it is by no means the only module that is important. The reason why this module occupies a larger space in this figure and has its details – the sub-modules – illustrated is that the CD+ representations that we will be using for illustrating the concepts involved in language communication reference these sub-modules in LCRC. The submodules in LCRC are PROBLEM SOLVING (PS), SIMULATION (SM), BUFFER(BF), and CONTROL(CT) modules.

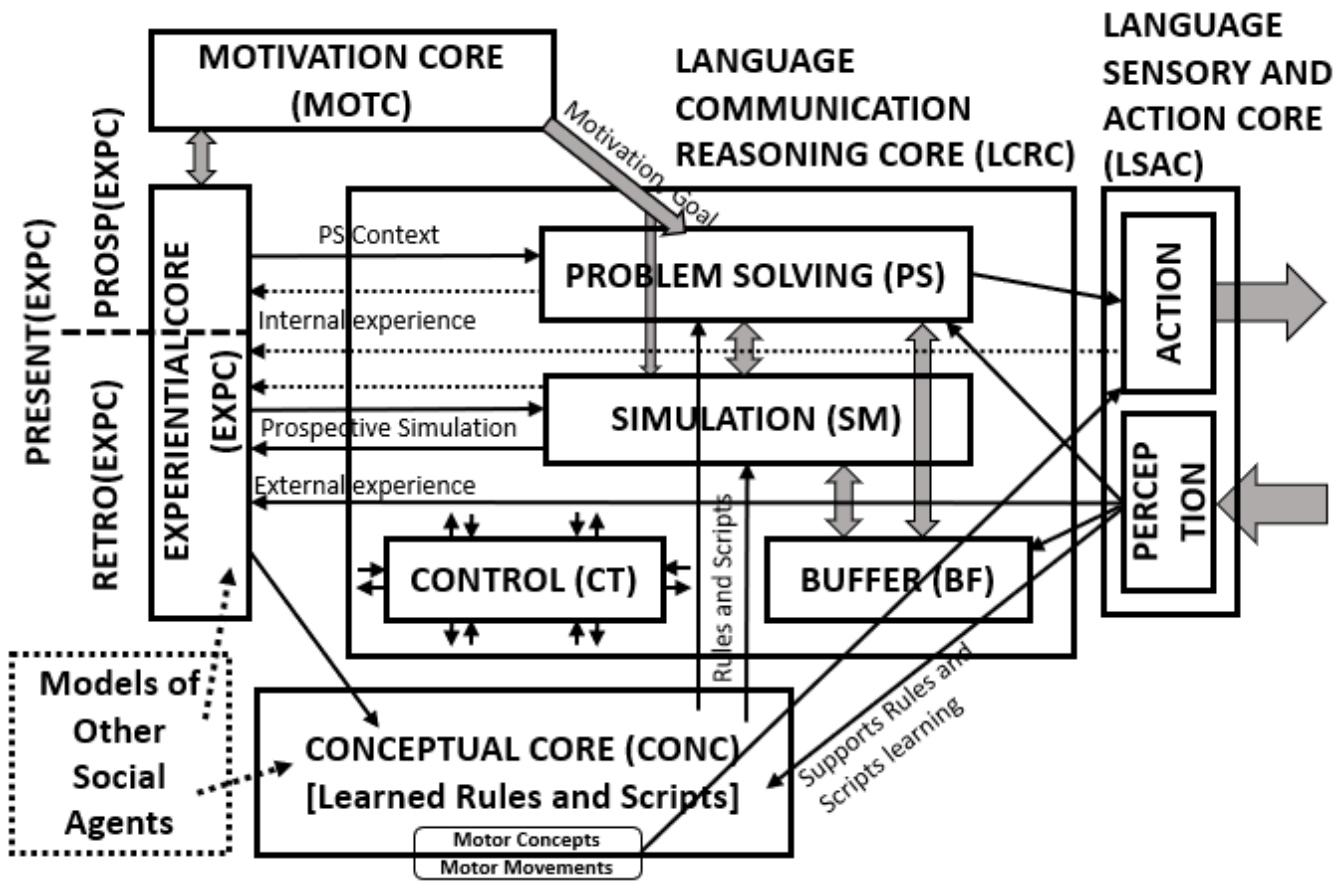

Fig. 2. The UGALRS architecture for an Intellient Autonomous System (IAS), a robot, or a social agent. Based on [12]. See text for explanation.

In the full UGALRS [12] there is a corresponding module, REASONING CORE (REAC) to LCRC, whose major input and output are the PERCEPTION and ACTION systems respectively (called the SENSORY AND ACTION CORE – SAAC) that are non-language related, but involve the usual perception and action processes. The basic idea behind UGALRS views the language communication process as similar to the usual “perception and action” processes related to vision and robotic limb action, but in the sphere of language. In the language sphere, “perception” is “language understanding,” and “action” is “language generation.”

These are done through the LANGUAGE SENSORY AND ACTION CORE (LSAC) here. To utter something, an intelligent autonomous system (IAS) would begin with some motivation, thus, the MOTIVATION CORE (MOTC). (On the REAC side, the MOTC would drive its reasoning to either understand the perceived information or to generate physical actions.) PS directs a problem-solving process to concoct an appropriate sentence to (hopefully) satisfy the motivation involved (like concocting an appropriate physical action through PS on the REAC side).

SM is used to simulate, based on some earlier learned language communication rules, what language responses from the intended recipient of the current utterance are expected. BF contains the concepts that are currently being operated on. CT directs and control all other modules.

The EXPERIENTIAL CORE (EXPC) records all experiences, linguistic or otherwise. Therefore, there is a path from the PERCEPTION module to EXPC. EXPC also records internal experiences such as what takes place in PS, SM, or what actions (utterances) are (were) emitted. EXPC also provides the context for PS, and directs prospective simulation in SM and receives its results. EXPC can be divided into 3 portions – the present (PRESENT(EXPC)), the past (RETRO(EXPC)) and the future (PROSP(EXPC)). RETRO(EXPC) and PROSP(EXPC) can be used to ground the concepts of past and future as illustrated in [12].

The CONCEPTUAL CORE (CONC) stores knowledge as rules and scripts represented in the form of CD+. Scripts are long causal sequences of events that pertain to certain knowledge complexes, as articulated in [15] and explained in the context of CD+ in [12]. In Fig. 2, it is shown that "Models of Other Social Agents" are stored in EXPC and CONC, in EXPC in the form of un-generalized instances, and in CONC, in the form of generalized knowledge. The dotted box of "Models of Other Social Agents" is not a functioning module of UGALRS, but serves to indicate the knowledge involved and where it is located in UGALRS. The sources of external knowledge for EXPC and CONC go through the PERCEPTION module in LSAC.

## IV. Representations for Lauguage Communication

Armed with the devices and constructs provided by UGALRS and CD+, in this section we illustrate their uses in representing the complex and intricate processes involved in language communication. It will be seen that between just a few sentences, many processes take place in the internal reasoning and problem solving modules of the utterer, whether it be human or robot (i.e., an IAS or a robotic social agent). These processes involve not only the conceptual, but also the motivational and affective, that can be represented by CD+ within UGALRS.

In the following, we will represent the internal conceptual, motivational, and affective processes in both Person and Robot using CD+. Even though the primary purpose of AI is in investigating the computational and representational processes in IAS (robots), there are two purposes in elucidating similar processes in the Person involved as well. First, Person can be another robot engaging in natural language communication with the first robot. Second, a robot or IAS can model the "mental" processes of another person or robot as well, which is the block indicated in the bottom left corner of Fig. 2. Therefore, in the following, we elucidate the processes taking place in the person as well as the robot.

### A. *Motivation and Sentence Concoction/Generation*

Before an utterance is made, the utterer must begin with an idea. This idea could be just a thought to be shared, or a want to be conveyed. Suppose a person (Person) thinks of asking a robot (Robot) to bring her a tool, Tool(X), from the table. This would be her "WANT," which if the robot could succeed in listening to her and satisfying it, she would be Pleased.

At the very top part of Fig. 3, this is represented after the same fashion as the representation of Fig. 1(b). Firstly, here are SAs associated with Person and Robot, and SAs such as these will be omitted in subsequent figures to avoid clutter. The PTRANS concept is Physical TRANSfer, used and explained in [13] and [12]. There is a "from" location (Loc(Table)) and a "to" location (Loc(Person)) and a Direction (D) of transfer. So, in this case, Person conceptualizes (CONCP) that if Robot were to PTRANS Tool(X) from Table to her, she would be Pleased. "I" on the right most side of the PTRANS representation represent the "Instrument" that Robot might use for this purpose, such as using its legs to propel itself along the ground. The PTRANS process involves a series of steps. Suppose Robot is currently next to Person and Table is some distance away, Robot would first turn its body and face Table, viewing from far to see that Tool(X) is on Table, then mentally, through a PS process, plot a path to Table, and after executing, say, a pickup action, bring Tool(X) to Person. This sequence of events is first worked out in Robot's REAC module (i.e., the usual non-language related problem solving and planning process). This is the HOW in the CD+ Elaboration (CD+E) pointed to from PTRANS.

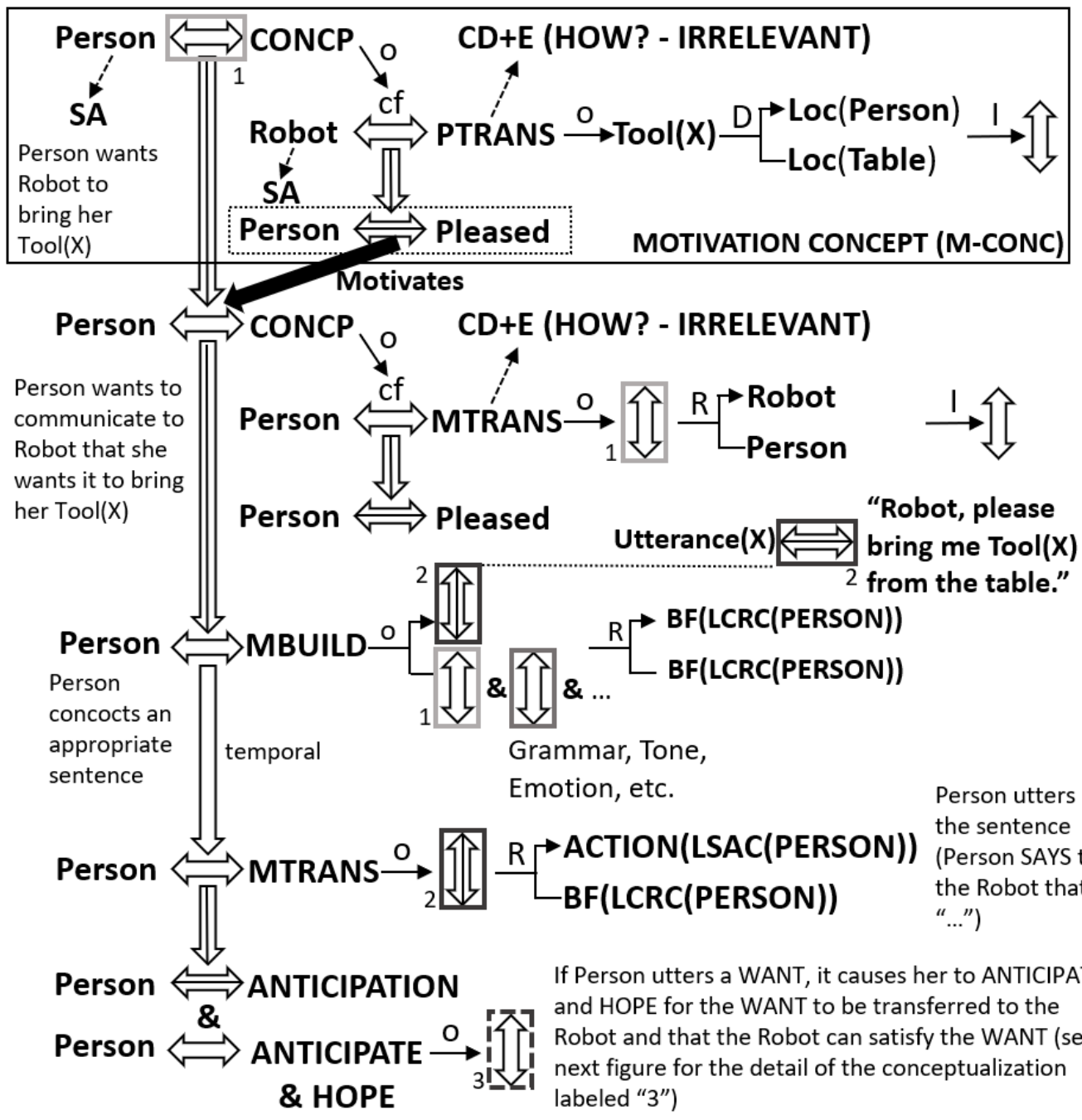

Fig. 3. The conceptual, motivational, and affective processes involved in the utterance "Robot, please bring me Tool(X) from the table." See text for explanation. The black "Motivates" arrow is not part of the representation, but an indication of the source of the causal link involved.

The desired Pleased state in the WANT conceptualization is a motivational force that propels Person to proceed to look for solutions to realize the Pleased state. Pleased is a basic and ground level emotion, as discussed in [12]

Now, as discussed in [12], when a WANT is conceptualized, there may or may not be a solution to satisfy the object of the WANT. Therefore, the exact "HOW" is "IRRELEVANT" in the concept of WANT. Hence, there could be a situation in which "I want to get rich but I can't." If a solution exists, then the concept of CAN comes into play. So, if "Robot CAN bring Tool(X) from Table to Person," then it means a solution exists. This representation of CAN is given in [12] and will be shown in Fig.

4. The entire CONCP encased in a box is called a MOTIVATION CONCEPT (M-CONC).

Next, having the WANT of a certain event (namely the Robot bringing Tool(X) to the Person), Person then WANTs to communicate the concept that she WANTs this certain event to happen to Robot. There is a general rule that says if an IAS (a human is a natural IAS while a robot is an artificial one) wants something, a motivation to achieve the state of "Pleased" will drive the IAS to do one of three things: 1. Carry out a planning or problem solving process to achieve the state of Pleased by herself or itself; 2. Request the help of someone else to do so; and 3. Command a servile agent to do so. This is encoded as the first causal link near the top of the figure. Now, note that at the very top of the figure, the WANT conceptualization (the topmost double arrow) is encased in a gray box with a label "1." This entire conceptualization "1", that the Person wants Robot to do something, is now the object of an MTRANS (Mental TRANSfer) process intended to "mentally" transfer the WANT conceptualization "1" from Person to Robot, that will make Person Pleased. (I.e., Person WANTs her WANT, currently in her mind, to be MTRANS to Robot's mind. In computational terms, "mind" is simply the internal memory and processing mechanisms of the human or robot involved.)

In order to realize this communication, Person concocts a sentence by Mentally BUILDing (MBUILDing) the sentence from considering the conceptualization involved (labeled "1"), together with the grammar of the language involved, the intended tone of the sentence, the emotional state of Person, etc. (the intended tone is dependent on the existing context of communication). The sentence constructed is "Robot, please bring me Tool(X) from the table," as shown in the figure as conceptualization "2". (The fact that Person WANTs Robot to do something is not explicitly stated in the sentence, but it is implied. Person could also have stated more explicitly, "Robot, I want you to bring me Tool(X) from the table.) The MBUILD conept is discussed in [13] and [12]. The MBUILDing of the sentence takes place in BF(LCRC(PERSON)). The precise process of converting an internal meaning representation in CD+ form to a grammatical surface sentence for communication is relegated to future work.

After the sentence is concocted in BF(LCRC(PERSON)), it it MTRANSed to ACTION(LSAC(PERSON)) to be emitted as an utterance. The link between MBUILD and MTRANS is a "temporal" one, not a causal one, as the second step simply follows the first step as part of the process (*temporal* links are indicated as a thickened arrow without a line running down its middle).

When a sentence, whether one that is a command or request, or just a factual statement, is uttered toward a recipient, a state of ANTICIPATION is entered (which is part of the implication of command or request) and the utterer then ANTICIPATEs something, as shown in the bottom of Fig. 3. This is the first affective state that emerges in the present communication process and will be discussed in detail in the next section. This ANTICIPATION is accompanied by HOPE as the prospect is positive [24].

### B. *Affective States and Illocutionary Forces*

In the non-linguistic sphere, when a certain action is emitted by an IAS, it is expected to cause certain effects. Similarly, in the linguistic sphere, an emitted utterance is expected to cause some effects. This has been investigated by speech act theorists [25]. If an utterance is meant to communicate certain information to the recipient, there may be no immediate overt actions or responses expected in the recipient, but the information conveyed may cause future actions or responses, or in the least, it may cause certain changes in the beliefs of the recipient. If the utterance is in the form of a command or request, immediate actions and responses are expected. The utterance is said to have an "illocutionary force." [25]

At the bottom of Fig. 3 we show that Person enters am affective state of ANTICIPATION and she ANTICIPATEs something. In Fig. 4(a) we show the functional representation of ANTICIPATE, an action that accompanies the affective state ANTICIPATION. If an Agent ANTICIPATEs a certain conceptualization, she conceptualizes that the conceptualization involved will happen in the future. It is shown in Fig. 4(a) that the object of Agent's CONCP is labeled with an "f", which means it resides in the prospective part of the EXPC (Fig. 2), PROSP(EXPC). This formulation of an affective state and its associated causal consequences is in consonant with the cognitive appraisal theory of emotion [24]. The same approach will be adopted with the other affective states in subsequent discussions.

Specifically, in the situation depicted in Fig. 3 in which Person asks Robot to bring her Tool(X) from Table, she ANTICIPATEs both the facts that "Robot WANTs to PTRANS Tool(X) from Table to Person so that Person is Pleased" and "Robot CAN PTRANS Tool(X) from Table to Person so that Person is Pleased," as shown in Fig. 4(b).

First, let us consider the representation for "Robot WANTs to PTRANS Tool(X) from Table to Person so that Person is Pleased." This is a transfer of what Person WANTs to what Robot WANTs. Now, for Person to reasonably assume that Robot would WANT to Please her, it must be assumed that Robot has either a SERVILE or an ALTRUISTIC attitude. In situations in which Robot or other recipient(s) of the utterance is REBELLIOUS or UNCOOPERATIVE, then this situation is not obtained. In Fig. 5 we depict that Robot is indeed SERVILE or ALTRUISTIC and hence in Fig. 4(b) Person ANTICIPATES that "Robot WANTs to PTRANS Tool(X) from Table to Person so that Person is Pleased." Hence, Robot being Pleased is caused by Person being Pleased. CD+ can be used to represent the connections between attitudes such as being SERVILE, ALTRUISTIC, COOPERATIVE, REBELLIOUS, or UNCOOPERATIVE and whether the entity/IAS involves WANTs to do certain things. The details of these are left to future work.

The locus of the illocutionary force is this. In any IAS, ultimately it will do whatever Pleases it. The arrow labeled PS shows the flow of Robot's actions: in order to please itself, it has to please Person, and in order to please Person, it has to PTRANS Tool(X) from Table to Person, if it CAN.

As mentioned above, there is a difference between WANT and CAN [12]. The primary difference is that WANTing something to happen (e.g., PTRANSing something from one place to another) does not imply that a solution exists for the thing to happen, but CAN implies that the solution exists. Therefore, there could be a situation that "I want to go from here to there but I can't". Hence, the representation for "Robot CAN PTRANS Tool(X) from Table to Person so that Person is Pleased" shown in Fig. 4(b) is that PS(REAC(ROBOT)) returns a solution (Solution(X)) for Robot to PTRANS Tool(X) from Table to Person. (EXTRANS stands for EXistential TRANSformation in which something goes from non-existence to existence or vice versa – which is used to represent the existence of a Solution(X) – see [12])

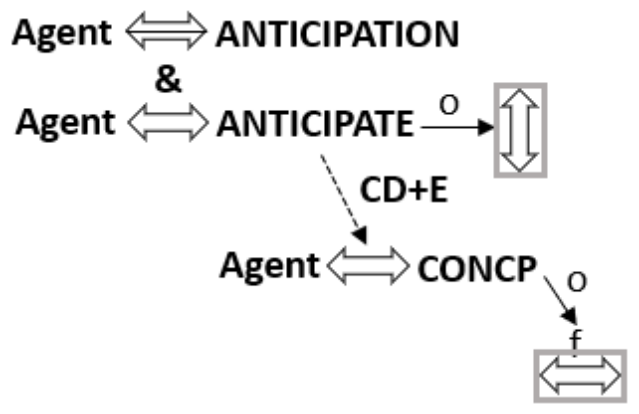

(a)

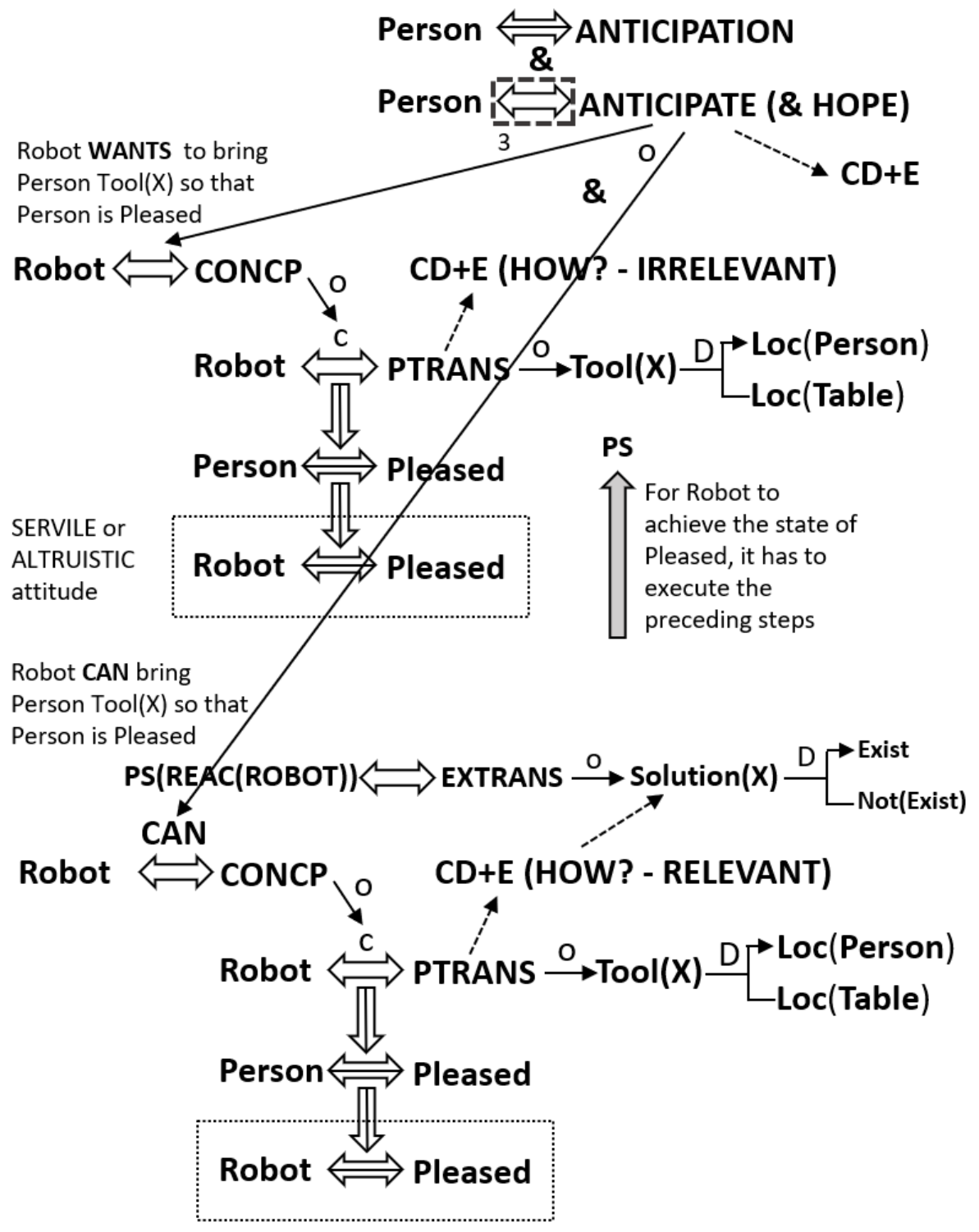

(b)

Fig. 4. (a) The general representation of ANTICIPATION. There is a *state* of ANTICIPATION, and an *object* referred to by the ANTICIPATE action. (b) The specific case of ANTICIPATION at the bottom of Fig. 3.

*C. Sentence Understanding, Actions, and Affective States*

Now that the utterance has been made and presumably received by Robot, the first step of the process on Robot's side is to MTRANS the received utterance into BF(LCRC(ROBOT)) for further processing as shown in Fig. 5. This causes Robot to MBUILD the conceptualization corresponding to the received utterance, based on the grammar of the language, the tone present in the sentence, the current emotional state of Robot and the perceived emotional state of Person, etc. This MBUILDed conceptualization is labeled "1", which is the same as conceptualization "1" in Fig. 3. Assuming Robot has either a SERVILE, COOPERATIVE, or ALTRUISTIC attitude, this causes it to create the next conceptualization capturing the fact that Robot will be Pleased if Person is Pleased due to Robot carrying out a certain task. This then motivates Robot to seek a solution to conceptualization "4", which is "Robot PTRANS Tool(X) from Table to Person." This conceptualization is MTRANS from BF(REAC(ROBOT)) to PS(REAC(ROBOT)).

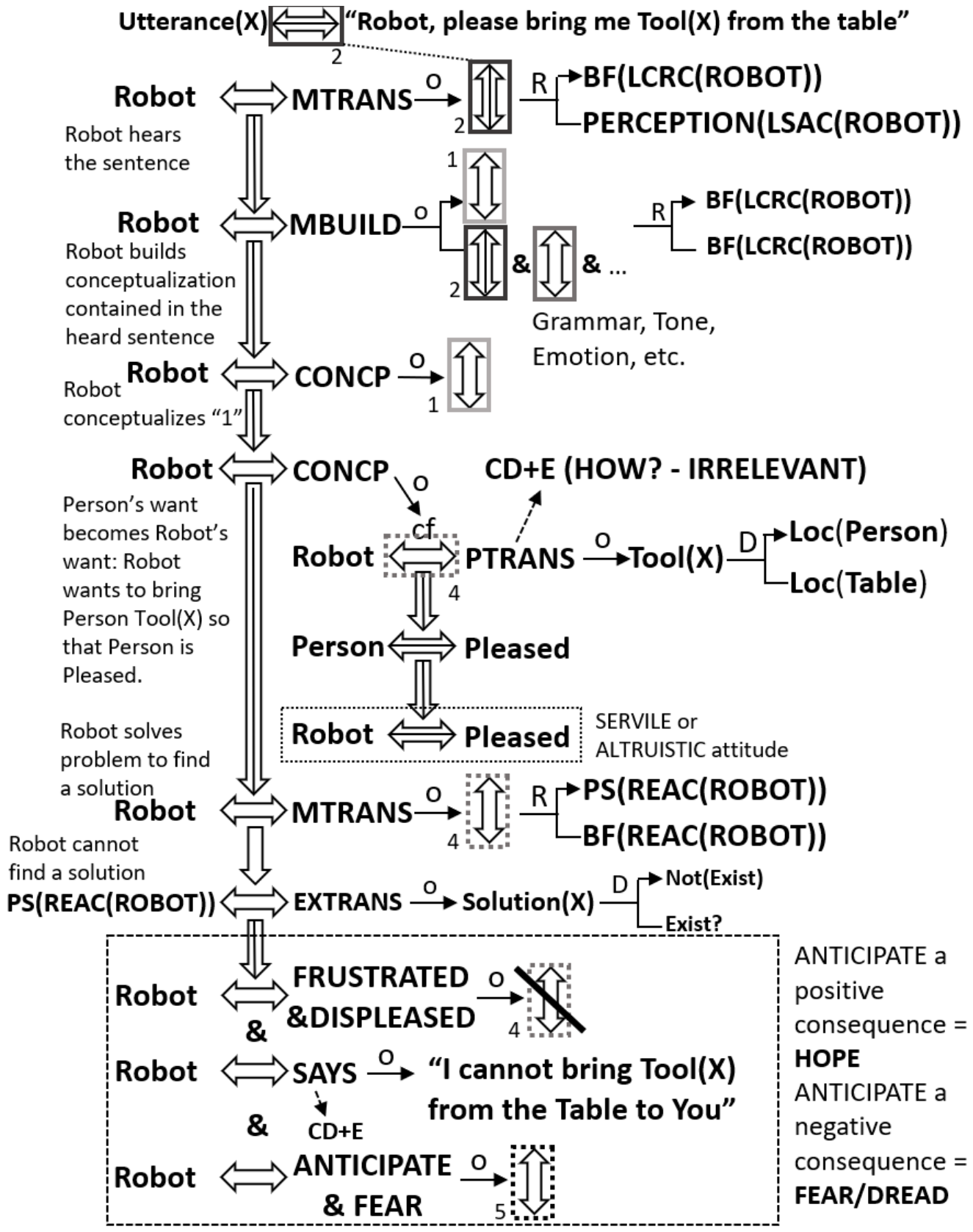

Fig. 5. Robot's internal processing in response to Person's utterance in Fig. 4. See text for explanation. Unlike in Fig. 3, ANTICIPATE here is accompanied by FEAR as it is anticipating anegative prospect [24].

Suppose PS(REAC(ROBOT)) *cannot* find a solution subsequently. This situation is represented in CD+ using EXTRANS showing that a solution does not exist (see [12] for the concept of CANNOT). It causes Robot to enter states of FRUSTRATED, DISPLEASED and FEAR (unlike for the case of ANTICIPATION in Figs. 3 and 4(a), these are not shown in Fig. 5 to avoid clutter) and it is also FRUSTRATED, DISPLEASED, and FEARful *about* the un-attainment of conceptualization "4" as shown in Fig. 5 ("un-attainment' is represented as a slash across the conceptualization and is directly related to the concept of CANNOT). FRUSTRATION and the other emotions can also arise if PS(REAC(ROBOT)) can find a solution but Robot cannot execute it due to other situations that are not anticipated in the PS process.

The state of FRUSTRATION always follows the situation when an Agent WANTs something but it cannot be obtained, as shown in Fig. 6. The state of Displeased also accompanies this based on a rule that states that if the object of a WANT is not achievable or satisfiable, the IAS involved will be Displeased. FEAR comes from the fact that Robot has a model of Person's negative response to the un-attainment of her WANT, and it is reflected in its ANTICIPATION of conceptualization "5", which is shown in Fig. 7 as "Person is DISAPPOINTED and DISPLEASED that conceptualization "4" is not attainable." FEAR is the anticipation of a negative consequence that may happen to the agent itself [24]. It is not shown here that a DISAPPOINTED and DISPLEASED Person toward Robot may take negative actions toward it in some way.

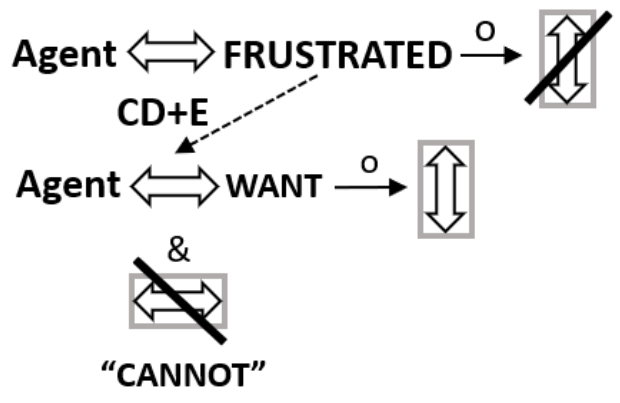

Fig. 6. The meaning and representation of FRUSTRATED.

Other than the consequences above, another response to not being able to find a solution includes Robot communicating this to Person by uttering "I cannot bring Tool(X) from the table to you" as shown in Fig. 5. The symbol "SAY" has a CD+E that is the same processes in Fig. 3 when Person concocts and utters a sentence, but we omit the detailed CD+E here. If instead Robot is able to find a solution, it would go ahead to carry out the task and say "Here is Tool(X)" when handing it to Person. What motives Robot to explain its failure is the communication rule that states: "*if others are displeased with your failure to do something on request or command, do communicate about it, including explaining the reason involved, because this will placate the other person, which in turn should reduce your own frustration, displeasure, and fear*." This entire rule could be stated in CD+ for the system to interpret and execute.

## *D. Continuing Communication*

Following Robot reporting that it is not able to bring Tool(X) to Person, Person enters the state of being DISAPPOINTED and DISPLEASED and the object of the DISAPPOINTment and DISPLEASure is the un-attainment of conceptualization "4", as shown in Fig. 7. Instead of just keeping quiet, which is a possible response on the part of Person if she is no longer concerned about the un-attainment of "4" or she is taking some time to ponder her response, a typical immediate response on the part of Person is to try and understand the *cause* of the un-attainment of "4." To this end, Person asks "Why can't you bring Tool(X) to me?" as shown in Fig. 7.

As the UGALRS and CD+ representational framework as articulate in [12] is a fully explainable framework, when in the problem solving process, PS(REAC(ROBOT)) fails to return a solution, because the steps of processing everywhere in UGALRS using CD+ representations are explicit, the cause(s) of the PS failure is easily identified. Hence, the Robot would response with "Because Tool(X) is not on the table."

What causes Robot to respond is the illocutionary force present in Person asking the Why question (i.e., it is in Robot's CONC, where general knowledge is stored – Fig. 2 – that Person would be Pleased if her Why question is answered to, and would be Displeased if this is no so. This knowledge is also represented in CD+ form in CONC. These are the representations of the illocutionary forces involved.)

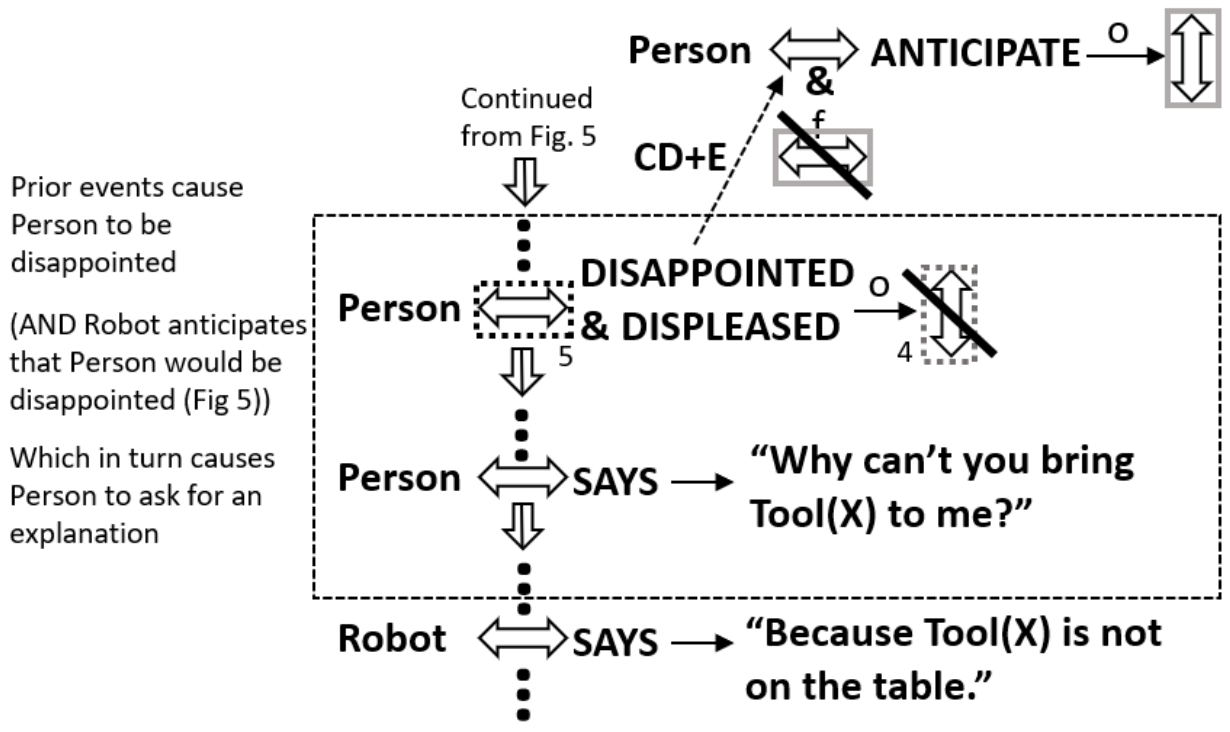

Fig. 7. Possible communication continued from Fig 5.

Robot may feel further RELIEVED from being FRUSTRATED, DISPLEASED, and FEARful after providing this explanation, because providing an explanation may cause Person to be more Pleased.

# V. Discussion and Conclusion

As illustrated in the foregoing discussions, many complex and intricate processes take place even between a small number of relatively simple utterances by an intelligent system, and these involve conceptual, motivational, and affective processes. We used the UGALRS and CD+ framework to elucidate some of these processes. For the sake of clarity and to avoid clutter, there are some processes that have been omitted in these diagrams, but some of them have been discussed in the texts, especially some of the communication rules underlying the generation of sentences. Future research could further develop in this direction to elucidate the explicit representations of the rules. Suffice it here to note that the CD+ representation is powerful enough to represent these rules and the situations under which they are triggered – i.e., the reasoning processes themselves are also representable using CD+, as has also been amply illustrated in [12].

Psychologists have identified up to 161 types of motivations in humans [26]. For robots, the number of motivations could be simpler and smaller in number [21]–[23]. However, for an IAS or robot to understand humans and hence be able to interact with them effectively, it has to have a model of the humans' motivations, as shown in Fig. 2 and discussed in this paper. Due to the limitation of space, in this paper we have only dealt with a small number of motivations. Future work certain calls for extension in this direction, and as demonstrated in [12], it is possible to do this within a UGALRS plus CD+ framework.

There is also a large number of emotions which are useful for characterizing and communicating about internal states in robots and humans than has been discussed here [10], [27], [28]. Ortony's cognitive appraisal theory of emotion [24], which lends itself conveniently for computational treatment, has been capitalized here to some extent, but there is a fully set that has not been dealt with in this paper. Therefore, a fuller set of emotions and a more complete treatment of affective processes would allow the IAS involved to deal with a more complete range of communicative scenarios.

Despite the fact that this paper only covers a subset of these vast conceptual, motivational, and affective spaces, its main contribution is to articulate a general framework of knowledge representation and processing to link these together and elucidate the respective functions they serve in the complex process underlying natural language communication.

Other important future work includes: 1. The transformations between the surface sentences illustrated in many places in this paper and their corresponding deep level "meaning" representations (Section IV(A)). This has also not been fully developed in Schank's original CD work [13]–[15]. The transformation must take into consideration grammar, tone, emotion, etc. 2. The roles, representations and causal consequences of various attitudes such as SERVILE, UNCOOPERATIVE, etc. (Section IV(B)). 3. Extension of the current framework and paradigm to cover a wider range of communication. 4. A computational implementation of the representation and processes involved. 5. The learning of the various representations illustrated in this paper.

That CD+ is a computationally viable representational scheme for this specific situation and other more general domains, even though a computational implementation is not reported in this paper, is reflected in the fact that in the original work of Schank and his associates [13]–[15], it has already been demonstrated that a computational implementation of CD could handle natural language question-answering and communication processes that benefit from the deep meaning representations of CD. It follows then that the enhanced form of CD, the CD+, as promulgated in [12] and in this paper, is also computationally implementable.

The learning of the representations discussed in this paper is also of paramount importance, as any system that cannot learn cannot be scaled up and is not viable as a practical system. Ho [12] discusses how learning could be done in the framework of CD+. However, it is also important to understand the kind of representations that are needed for intelligent processes, in this case, language communication processes, before we understand what it is that is to be learned. This paper hence contributes to the elucidation of the intricate and complex conceptual, motivational, and affective processes involved in natural language communication between social agents, which, when appropriately extended, would hopefully bring about a fuller characterization of language communication in general.